\documentclass[10pt,twocolumn,letterpaper]{article}

\usepackage{cvpr}
\usepackage{times}
\usepackage{epsfig}
\usepackage{graphicx}
\usepackage{amsmath}
\usepackage{authblk}

\usepackage{subcaption}
\usepackage{siunitx}
\usepackage{booktabs}
\usepackage{array}
\usepackage{gensymb}

\usepackage[pagebackref=true,breaklinks=true,letterpaper=true,colorlinks,bookmarks=false]{hyperref}

\cvprfinalcopy 

\def\cvprPaperID{9} 

\ifcvprfinal\fi
\begin{document}

\title{Registration made easy -- standalone orthopedic navigation with HoloLens}
\author[1,2]{Florentin Liebmann}
\author[1,3]{Simon Roner}
\author[1,2]{Marco von Atzigen}
\author[3]{Florian Wanivenhaus}
\author[ ]{Caroline Neuhaus}
\author[3]{Jos\'e Spirig}
\author[5,6]{Davide Scaramuzza}
\author[7]{Reto Sutter}
\author[2,3]{Jess Snedeker}
\author[3]{Mazda Farshad}
\author[1]{Philipp F\"urnstahl}

\affil[1]{ Computer Assisted Research \& Development, Balgrist University Hospital}
\affil[2]{ Laboratory for Orthopaedic Biomechanics, ETH Zurich}
\affil[3]{ Orthopaedic Department, Balgrist University Hospital, University of Zurich}
\affil[5]{ Department of Informatics, University of Zurich}
\affil[6]{ Department of Neuroinformatics, University of Zurich and ETH Zurich}
\affil[7]{ Radiology Department, Balgrist University Hospital, University of Zurich}

\maketitle
\thispagestyle{empty}

\begin{abstract}
	In surgical navigation, finding correspondence between preoperative plan and intraoperative anatomy, the so-called registration task, is imperative. One promising approach is to intraoperatively digitize anatomy and register it with the preoperative plan. State-of-the-art commercial navigation systems implement such approaches for pedicle screw placement in spinal fusion surgery. Although these systems improve surgical accuracy, they are not gold standard in clinical practice. Besides economical reasons, this may be due to their difficult integration into clinical workflows and unintuitive navigation feedback. Augmented Reality has the potential to overcome these limitations. Consequently, we propose a surgical navigation approach comprising intraoperative surface digitization for registration and intuitive holographic navigation for pedicle screw placement that runs entirely on the Microsoft HoloLens. Preliminary results from phantom experiments suggest that the method may meet clinical accuracy requirements.
\end{abstract}
%
%
\section{Introduction}
Finding correspondence between preoperative surgical plan and intraoperative anatomy is one of the core tasks in surgical navigation. This process is referred to as registration. Markelj \etal \cite{markelj2012review} give a good overview on clinical imaging modalities and the resulting registration possibilities. Preoperative data are 3D computed tomography (CT) or magnetic resonance images. In orthopedic surgery, common 2D imaging modalities are ultrasound (US), fluoroscopy or optical images. Cone-beam CT (CBCT), US and reconstruction with optical systems are 3D data sources. Consequently, intraoperative registrations are either 3D/2D or 3D/3D. 3D/3D is further subdivided into \textit{image-to-image} and \textit{image-to-patient} (when reconstruction with optical systems is employed).

In terms of 3D/2D registration, the most widely used approach involves digitally reconstructed radiographs (DRR), i.e. simulated 2D views generated from preoperative 3D data. The DRR are compared to intraoperative 2D fluoroscopic images using a similarity measure which can be optimized in an iterative fashion. However, surgical tools, implants and resected soft tissue which are usually not present in preoperative images make this approach challenging \cite{de20163d}. DRR generation is computationally expensive and a balance between robustness and computational cost has to be found \cite{miao2018dilated}. Moreover, a calibrated intraoperative setup is required.

Apart from a technical point of view, any intraoperative imaging that implies radiation exposure for patient and OR personnel is considered critically (e.g. \cite{narain2017radiation}). This adversely affects the use of intraoperative fluoroscopy and especially CBCT. 2D and 3D US are radiation-free alternatives. Nevertheless, they are not well-suited for open surgery due to their reliance on a transportation medium (e.g. \cite{yan2012ultrasound}).

Summarizing above findings, radiation-free 3D/3D image-to-patient registration can be considered most promising. An exemplary method following this approach is \textit{Surface Matching}, in which a 3D reconstruction of the visible anatomy is achieved by sampling bone surface with a tracked pointer. It is implemented in state-of-the-art commercial navigation systems and is well-known for pedicle screw placement in spinal fusion surgery \cite{richter2005cervical,nottmeier2007timing,chiang2012computed}. Although such systems increase accuracy, they are not established as the clinical gold standard \cite{andress2018fly}. Besides high costs for acquisition and maintenance, the main reason may be that they cannot be integrated into clinical practice without effort and hinder seamless workflows (e.g. line of sight issues) \cite{victor2004image}. Furthermore, anatomical 3D models and navigation feedback are not superimposed in situ, but visualized on peripheral 2D monitors which makes it hard to comprehend the underlying 3D space in an intuitive way \cite{qian2017towards}.

Augmented Reality (AR) has the potential to overcome these limitations. It has been considered interesting for medical applications for more than a decade \cite{sielhorst2008advanced,navab2012first}, but only recent technological advancement allows for the development of computationally powerful, off-the-shelf optical see-through head mounted-displays \cite{qian2017comparison}. We proposed a radiation-free surgical navigation approach comprising intraoperative manual surface digitization for registration and intuitive holographic navigation for pedicle screw placement that runs entirely on the Microsoft HoloLens \cite{liebmann2019pedicle}. The methodology of our approach and a pre-clinical evaluation are presented together with another in-house HoloLens application related to spinal fusion surgery \cite{wanivenhaus2019augmented}.
%
%
\section{Methods}
The presented method consists of two main components: intraoperative registration and surgical navigation. Both rely on 6DoF marker tracking.
\subsection{Marker tracking and pose estimation}
\label{subsec:marker_tracking}
Marker tracking was implemented employing the two front-facing of the four environment tracking cameras that are accessible via Research Mode \cite{Microsoft2018Research}. Commercial, sterile fiducial markers (Clear Guide Medical, Baltimore MD, USA) were used. Their patterns originate from the AprilTags library \cite{Olson2011,Wang2016}. We refer to the official terminology of the HoloLens coordinate systems to describe the process: App-specified Coordinate System (ASCS), Camera Coordinate System (CCS), 3D Camera View Space (CVS) and 2D Camera Projection Space (CPS). CVS and CPS of the left and right camera are termed CVSL/CVSR and CPSL/CPSR, respectively. An exemplary transformation from CPS to CVS is denoted as $T_{CPS}^{CVS}$.

For each pair of images (left and right) with a detectable marker, its pose is derived as follows. Initial estimate values $C_1^L,\dots,C_4^L$ and $C_1^R,\dots,C_4^R$ of the four corners of the marker are detected in both images using the ArUco library \cite{Garrido2014,Garrido2016,Romero2018}. Due to the low resolution ($480\times640$ pixels) of the environmental cameras, each $C_i$ is passed to a dedicated Kalman filter with a constant velocity model \cite{Kalman1960,Bradski2000}.

The filtered corner estimates are transformed to CPS and extended by one dimension (unit plane: $z=1$) such that they can be expressed in CVS and further transformed to CCS. To perform triangulation, directional vectors $\vec{d}_i^L$ and $\vec{d}_i^R$ between each $T_{CPSL}^{CCS}C_i^L$ and $T_{CPSR}^{CCS}C_i^R$ and their respective camera centers are formed. Triangulation can be completed by finding the closest point $min(\vec{d}_i^L,\vec{d}_i^R)$ between each pair of directional vectors.

Given the new 3D estimates $min(\vec{d}_i^L,\vec{d}_i^R)$, the 6DoF marker pose is derived by incorporating prior knowledge about the marker geometry. As the true 3D position $gt_i$ of each corner point with respect to the marker center is known, the pose estimation problem can be reduced to finding a rigid transformation between the point pairs $min(\vec{d}_i^L,\vec{d}_i^R)$ and $gt_i$ in a least-square sense. The transformation results from applying the absolute orientation \cite{Horn1987}.
\subsection{Registration}
\label{subsec:registration}
The idea of our surface digitization approach is to establish correspondence between pre- and intraoperative anatomy without using intraoperative imaging.

\begin{figure}[htb]
	\begin{center}
		\begin{subfigure}[t]{0.325\linewidth}
			\includegraphics[width=\linewidth]{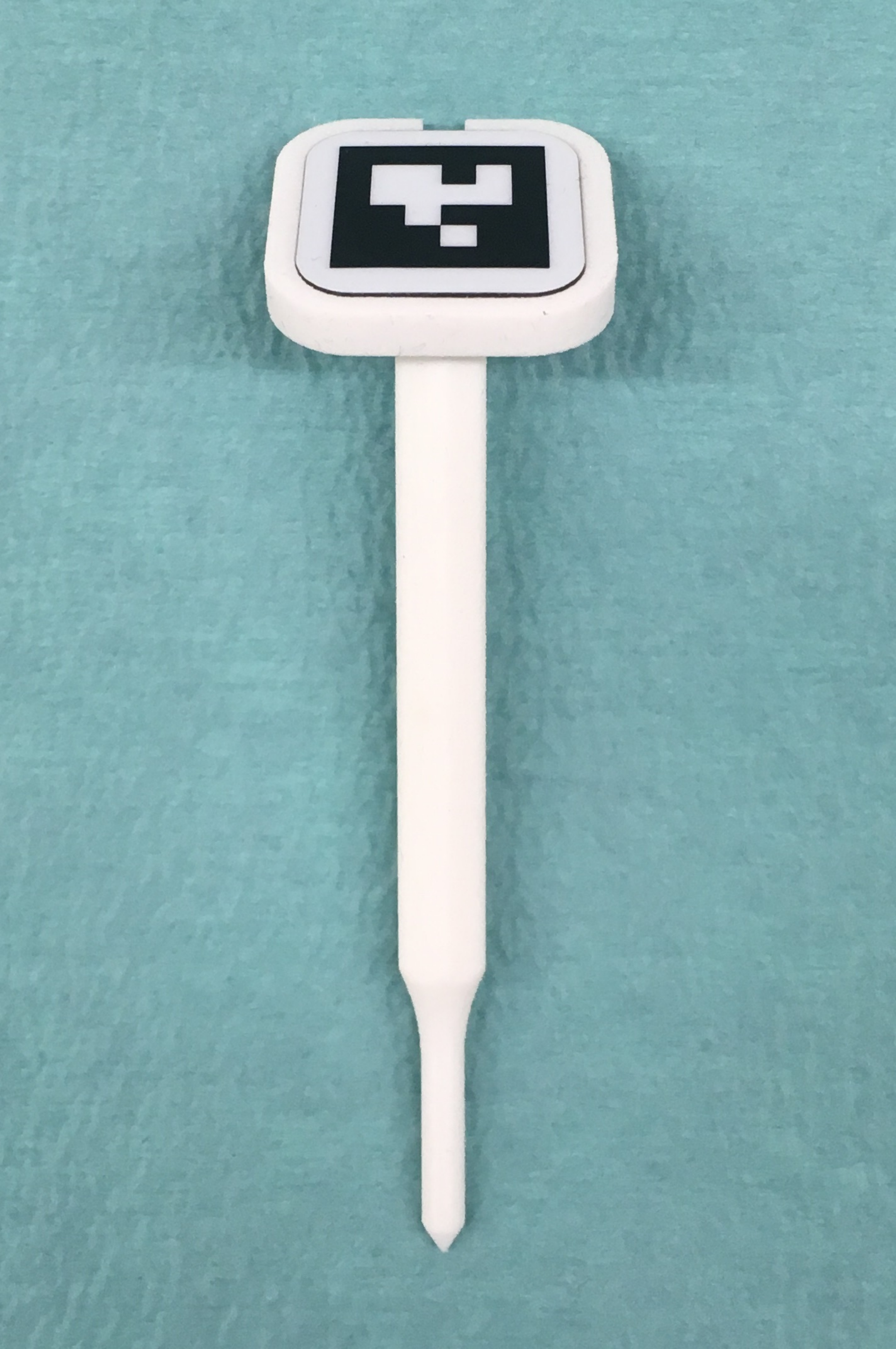}
			\caption{}
			\label{fig2a}
		\end{subfigure}
		\begin{subfigure}[t]{0.325\linewidth}
			\includegraphics[width=\linewidth]{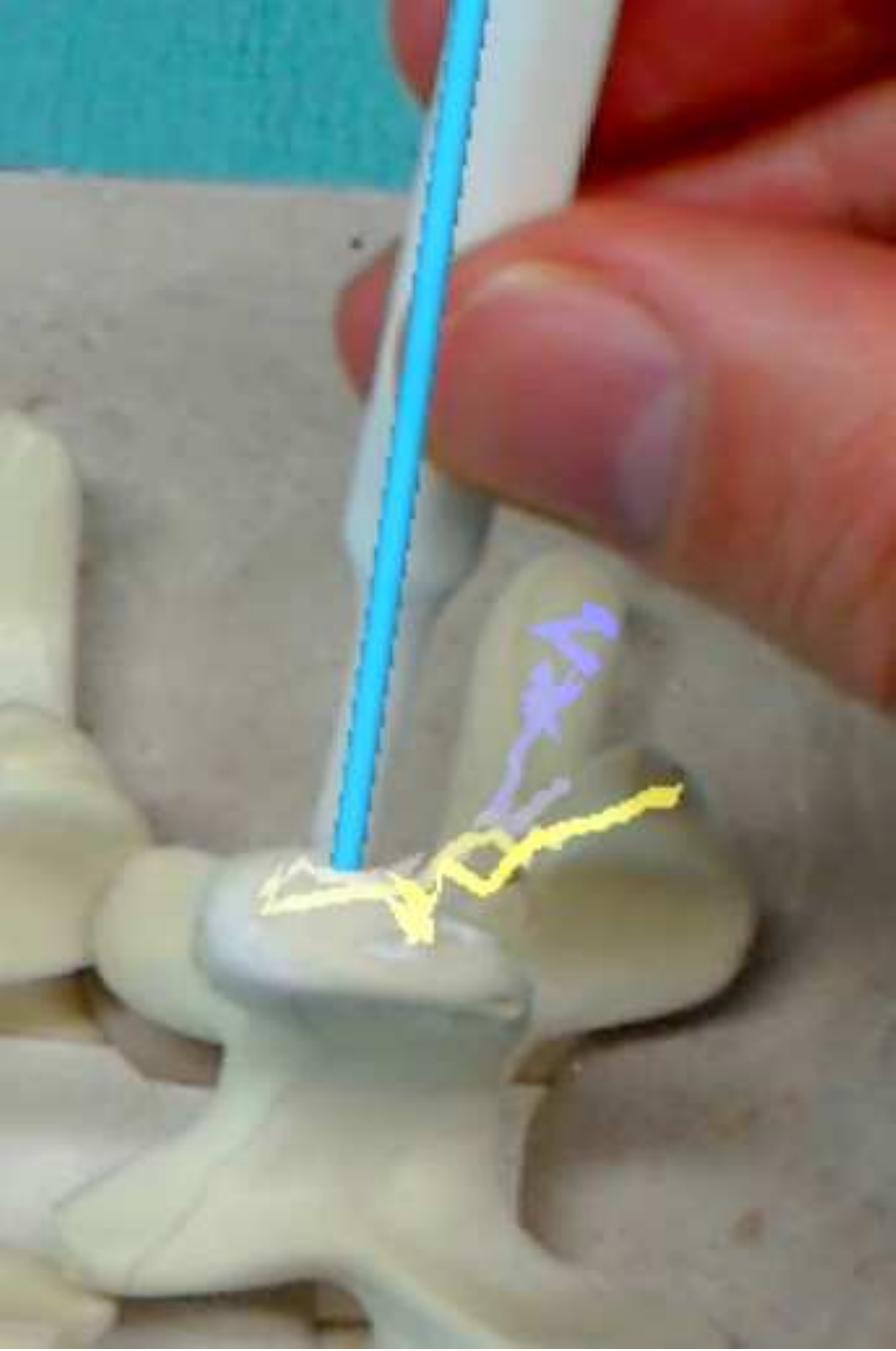}
			\caption{}
			\label{fig2c}
		\end{subfigure}
		\begin{subfigure}[t]{0.325\linewidth}
			\includegraphics[width=\linewidth]{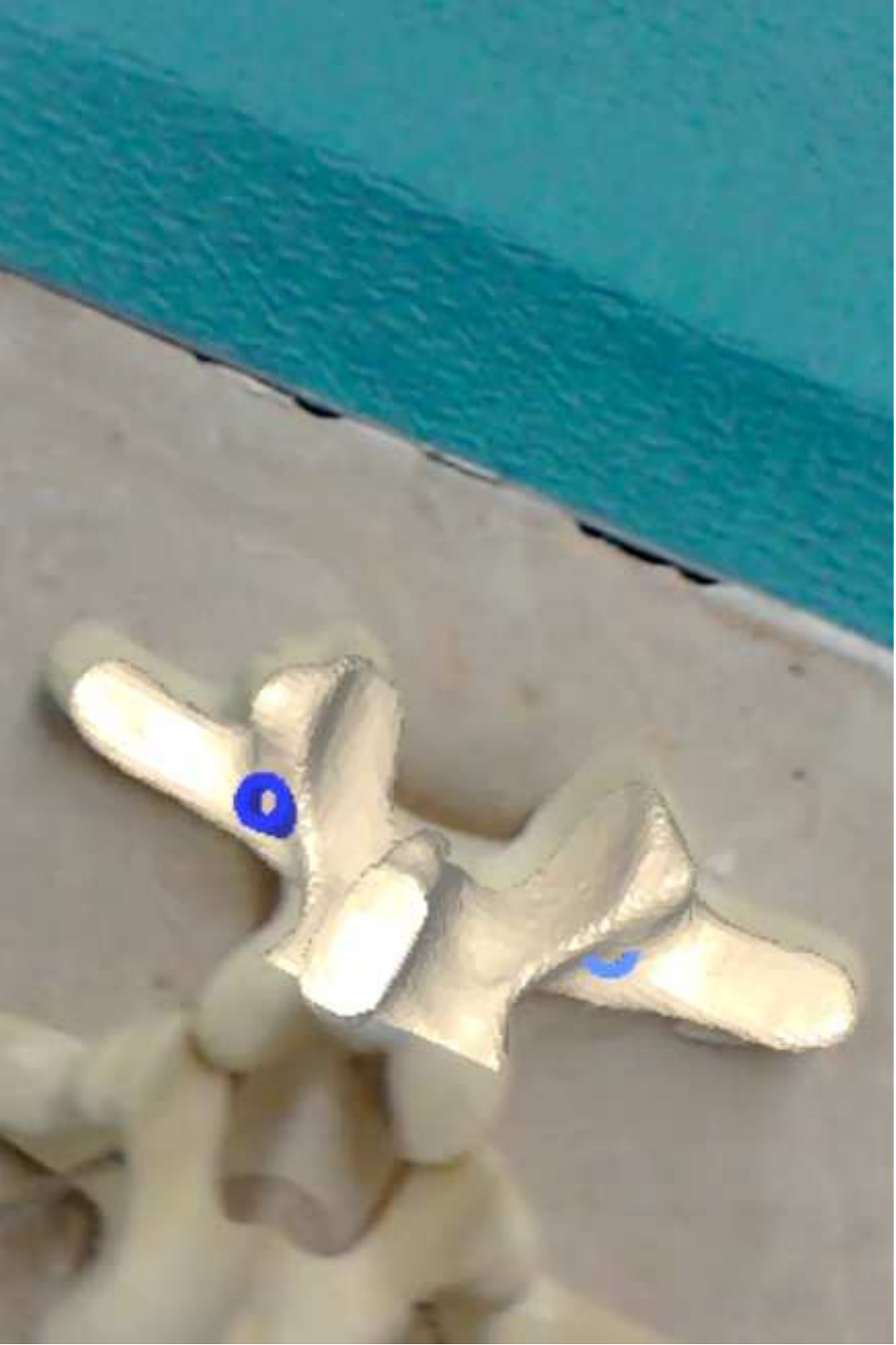}
			\caption{}
			\label{fig2d}
		\end{subfigure}
	\end{center}
	\caption{a) Custom-made pointing device. b) Augmented view of the surgeon during surface sampling. c) Vertebra overlay after registration (insertion points denoted in blue).
		\vspace{2mm}\newline\noindent
		\tiny
		\textit{Reprinted by permission from RightsLink: Springer International Journal of Computer Assisted Radiology and Surgery \cite{liebmann2019pedicle}, \textcopyright\space CARS (2019), advance online publication, 15 April 2019 (doi: \href{https://doi.org/10.1007/s11548-019-01973-7}{10.1007/s11548-019-01973-7}.IJCARS.)}
	}
	\label{fig2}
\end{figure}

For each vertebra, a sparse point cloud of relevant bone surface regions is collected by the surgeon. A custom-made pointing device (PD, Figure \ref{fig2a}) is used for surface acquisition. The PD consists of a notch, a handle and a tip. The tip is tapered in a way such that points can be reached at different angles without introducing an offset. The notch can be mounted with a previously described marker (Section \ref{subsec:marker_tracking}). The known geometry of the PD makes it straightforward to extrapolate from marker pose to tip position.

After application startup, the surgeon is asked to sample accessible surface regions of the vertebra in a specific pattern which was trained previously. To do so, the PD is moved along the anatomy while pressing down the button of the HoloLens clicker. The 3D tip position is recorded, as long as the button remains pressed. Sampled areas are visualized by a thin line connecting consecutively collected points (Figure \ref{fig2c}). Once the button is released, a voice command can be used to indicate whether the just collected region should be saved (``save'') or discarded (``delete''). Only the saved points are used for registration. When sufficient points have been sampled, a double click starts registration.

The intraoperatively collected point cloud $pc_{intra}$ is registered to the point cloud $pc_{pre}$ representing the points of the 3D model of the preoperative vertebra. In a preprocessing step, $pc_{pre}$ is trimmed by removing points which can definitely not be reached with the PD in a surgery.

\begin{figure}[htb]
	\begin{center}
		\begin{subfigure}[t]{0.49\linewidth}
			\includegraphics[width=\linewidth]{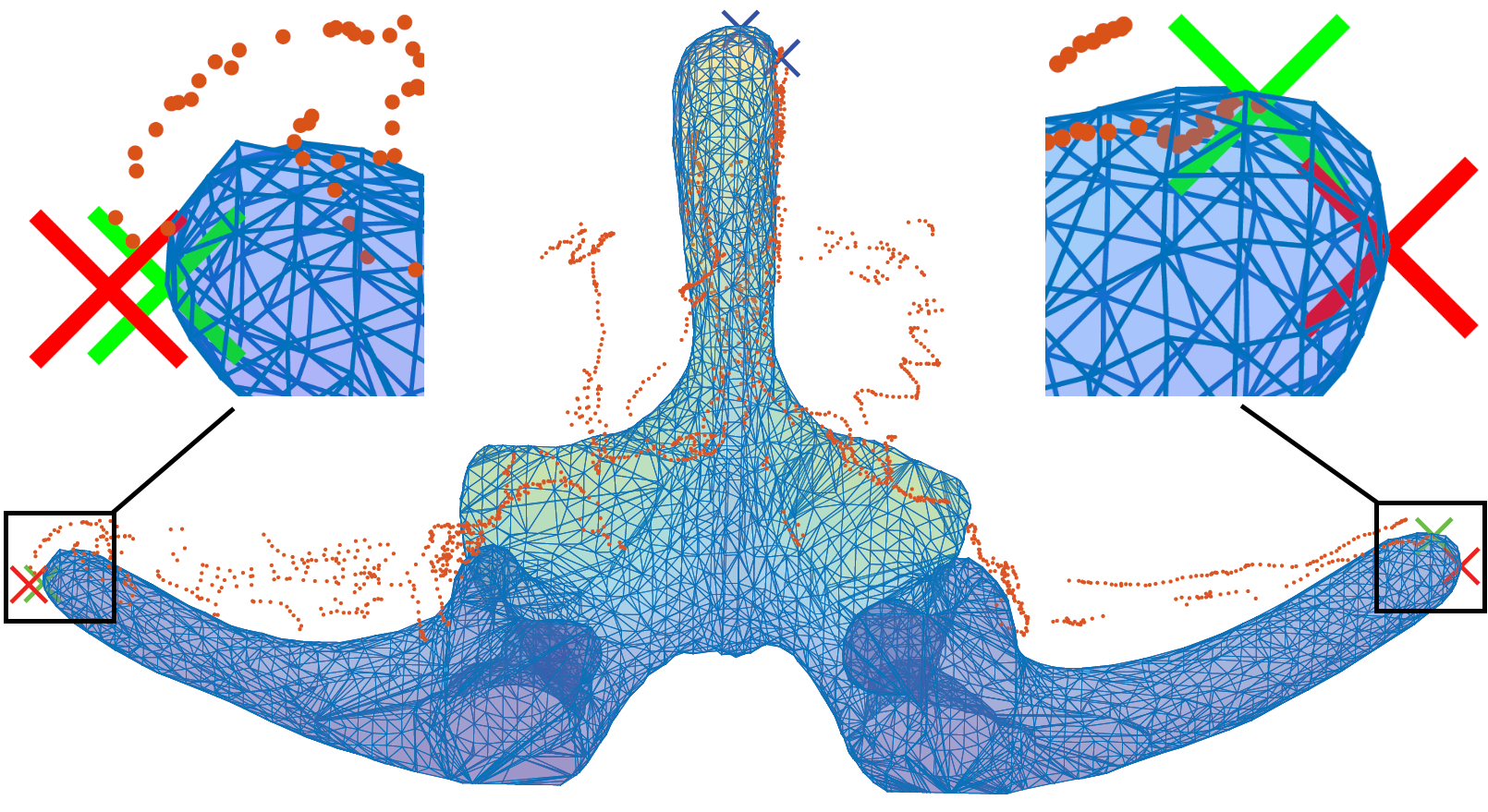}
			\caption{}
			\label{fig4a}
		\end{subfigure}
		\begin{subfigure}[t]{0.49\linewidth}
			\includegraphics[width=\linewidth]{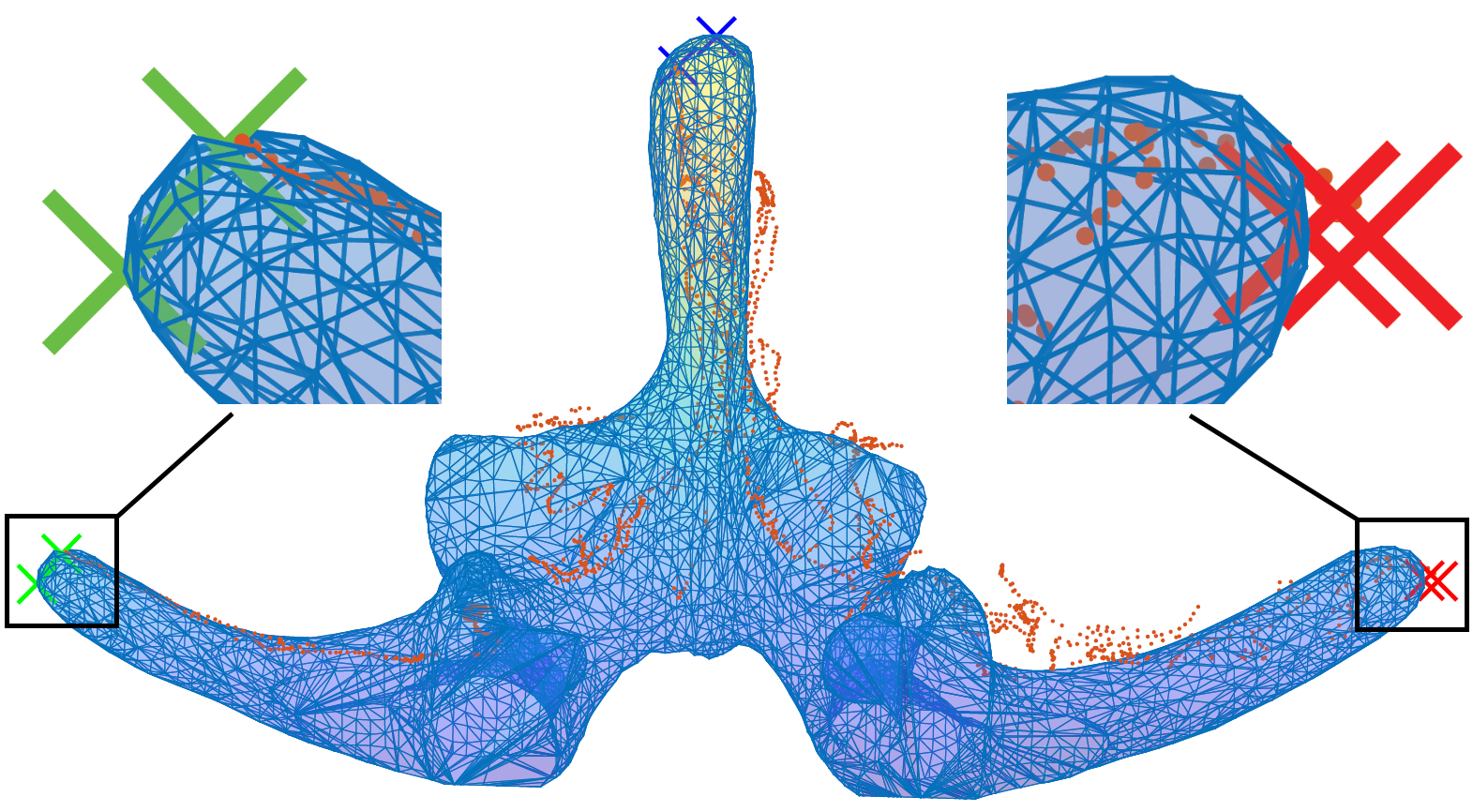}
			\caption{}
			\label{fig4b}
		\end{subfigure}
		\begin{subfigure}[t]{\linewidth}
			\includegraphics[width=\linewidth]{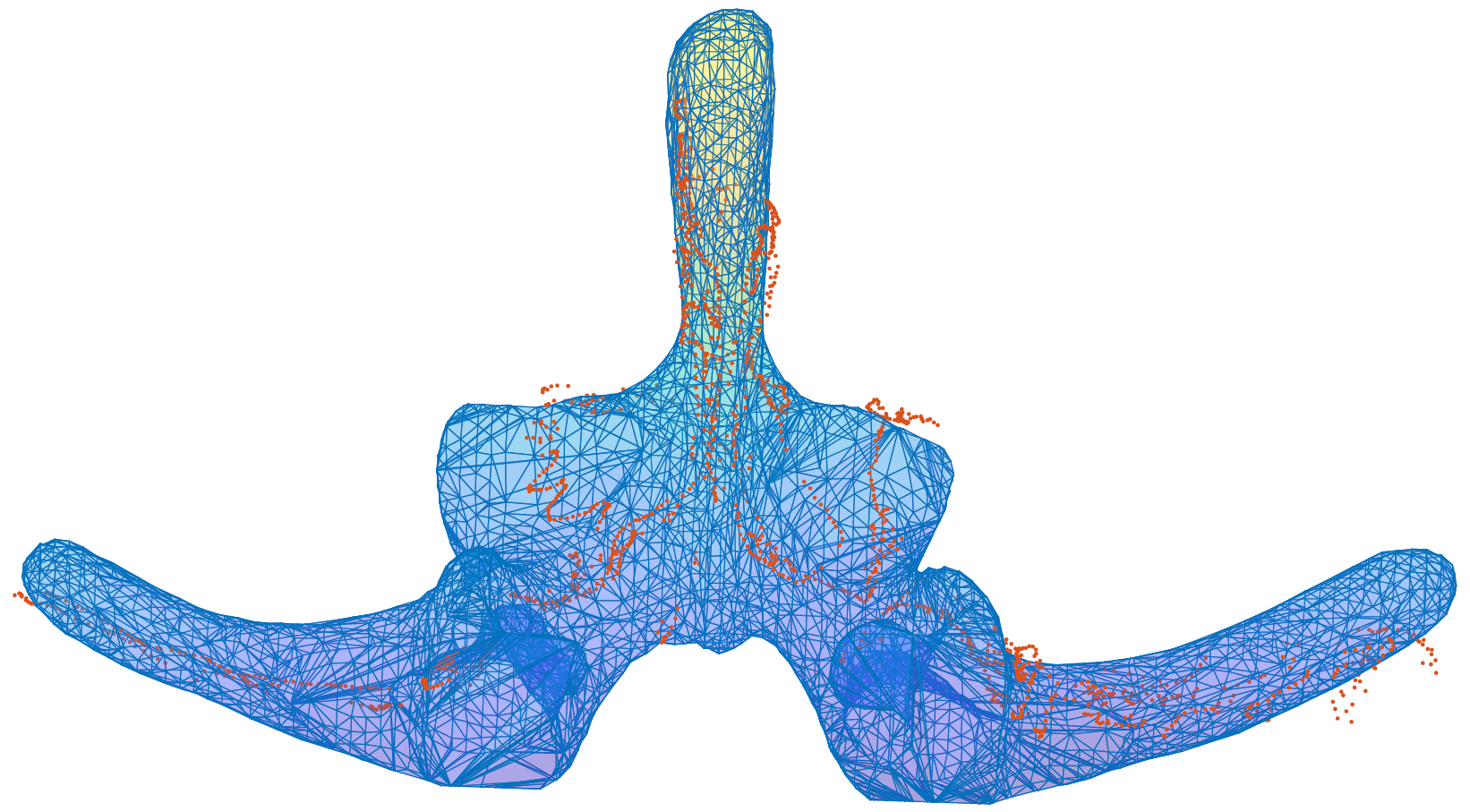}
			\caption{}
			\label{fig4c}
		\end{subfigure}
	\end{center}
	\caption{$pc_{pre}$ denotes the points of the 3D model. $pc_{intra}$ is shown in orange. The red, green and blue crosses in a) and b) represent the respective extreme points used for coarse alignment. a) Incorrect coarse alignment. b) Correct coarse alignment. c) The fine alignment resulting from b).
		\vspace{2mm}\newline\noindent
		\tiny
		\textit{Reprinted by permission from RightsLink: Springer International Journal of Computer Assisted Radiology and Surgery \cite{liebmann2019pedicle}, \textcopyright\space CARS (2019), advance online publication, 15 April 2019 (doi: \href{https://doi.org/10.1007/s11548-019-01973-7}{10.1007/s11548-019-01973-7}.IJCARS.)}
	}
	\label{fig4}
\end{figure}

The automated registration process comprises three steps: coarse registration, iterative closest point (ICP) based fine registration \cite{Besl1992} and result selection. Coarse registration is achieved by identifying three corresponding extreme points in each of the point clouds. To this end, a principle component analysis (PCA) \cite{Pearson1901}, implemented in ALGLIB (ALGLIB Project, Nizhny Novgorod, Russia), is performed on $pc_{intra}$, yielding the respective principle axes $pa_{1}^{intra}$, $pa_{2}^{intra}$ and $pa_{3}^{intra}$ ordered by decreasing magnitude. The three extreme points $e_{1}^{intra}$, $e_{2}^{intra}$ and $e_{3}^{intra}$ are determined using the dot product (Figures \ref{fig4a} and \ref{fig4b}):
\begin{equation}
\begin{aligned}
e_{1}^{intra}&=max(pa_{1}^{intra}\cdot p_{i}),\forall p_{i}\in pc_{intra}\\
e_{2}^{intra}&=min(pa_{1}^{intra}\cdot p_{i}),\forall p_{i}\in pc_{intra}\\
e_{3}^{intra}&=max(|pa_{2}^{intra}\cdot p_{i}|),\forall p_{i}\in pc_{intra}\\
\end{aligned}
\end{equation}
Correspondingly, the extreme points $e_{1}^{pre}$, $e_{2}^{pre}$ and $e_{3}^{pre}$ are calculated. Due to vertebra symmetry along $pa_{1}$, two possible coarse registration configurations must be evaluated (Figures \ref{fig4a} and \ref{fig4b}) and considered for the fine registration by applying absolute orientation \cite{Horn1987} to both point pair sets:
\begin{equation}
\begin{aligned}
\{(e_{1}^{intra},e_{1}^{pre})\},\{(e_{2}^{intra},e_{2}^{pre})\},\{(e_{3}^{intra},e_{3}^{pre})\}\\
\{(e_{1}^{intra},e_{2}^{pre})\},\{(e_{2}^{intra},e_{1}^{pre})\},\{(e_{3}^{intra},e_{3}^{pre})\}
\end{aligned}
\end{equation}
Afterwards, fine registration is performed using ICP on both configurations and the one with the smaller final RMSE is selected (Figure \ref{fig4c}). The result is shown to the surgeon by superimposing the preoperative 3D model with the intraoperative anatomy (Figure \ref{fig2d}) and is verified visually.
\subsection{Clinical application}
\label{subsec:surg_navigation}
In spinal fusion surgery, screws are inserted into the pedicles of pathological spine levels. The screw heads are then rigidly connected to each other by means of a rod \cite{harms1998posteriore}. Accurate screw insertion is crucial to avoid harming vital structure as nerves. Bending said rod can be complicated in complex deformity cases. Commonly, a wire is used in situ to form a template. The rod is then bent ex situ according to the wire using dedicated tools. This can be tedious process. To this end, we propose navigation methods for screw insertion and rod bending. Please note that in our study K-Wires instead of actual pedicle screws were inserted.

\begin{figure}[b]
	\begin{center}
		\begin{subfigure}[t]{0.325\linewidth}
			\includegraphics[width=\linewidth]{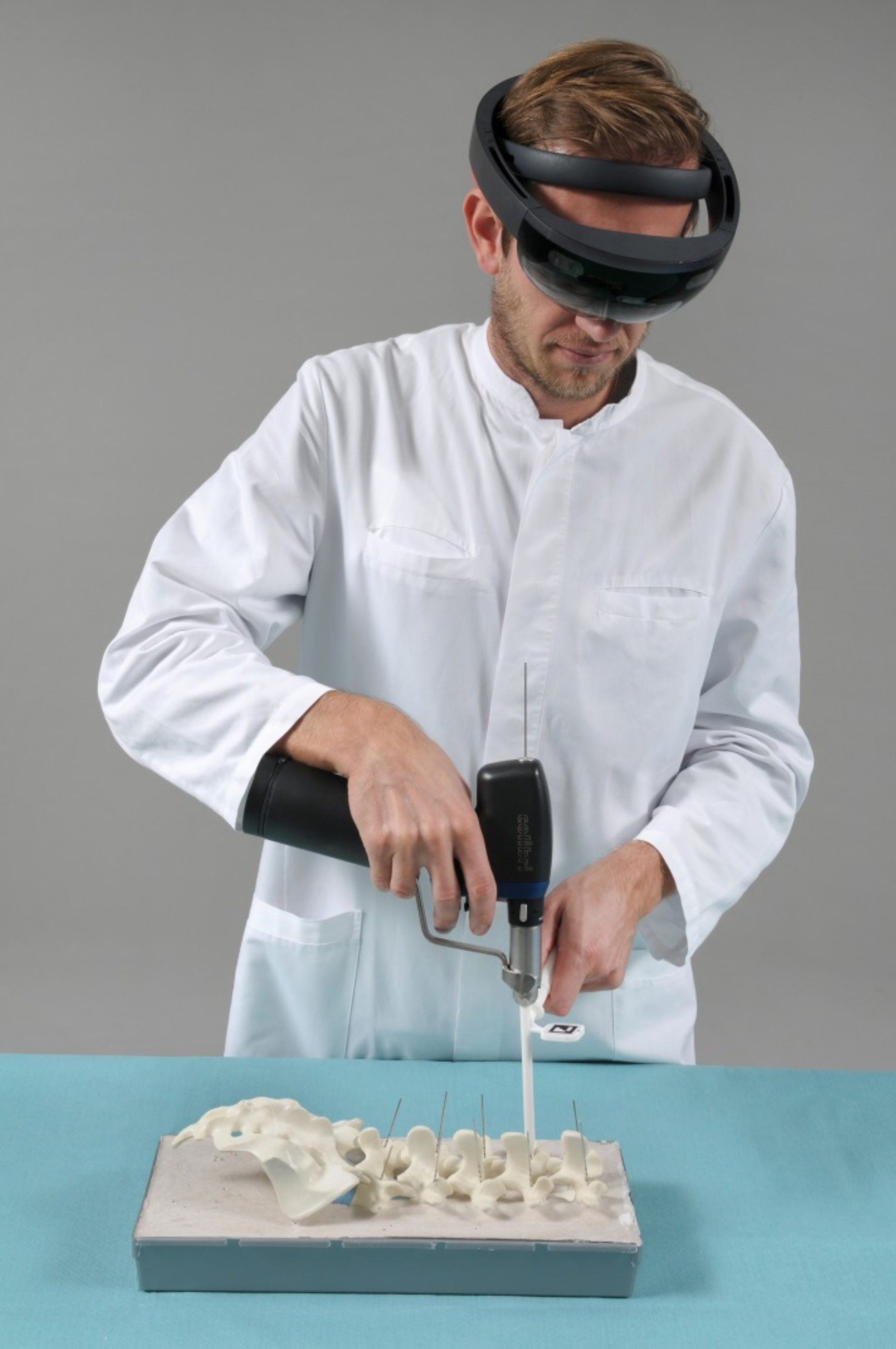}
			\caption{}
			\label{fig5b}
		\end{subfigure}
		\begin{subfigure}[t]{0.325\linewidth}
			\includegraphics[width=\linewidth]{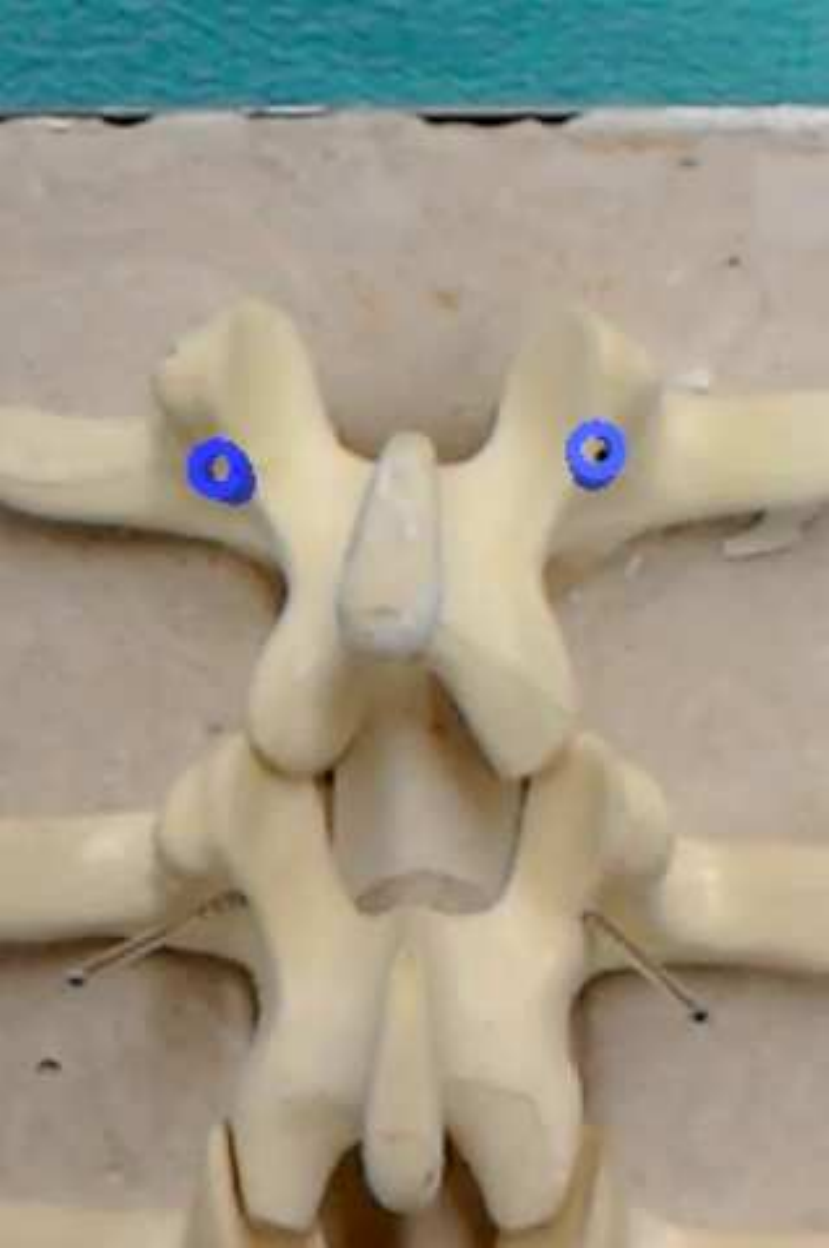}
			\caption{}
			\label{fig5c}
		\end{subfigure}
		\begin{subfigure}[t]{0.325\linewidth}
			\includegraphics[width=\linewidth]{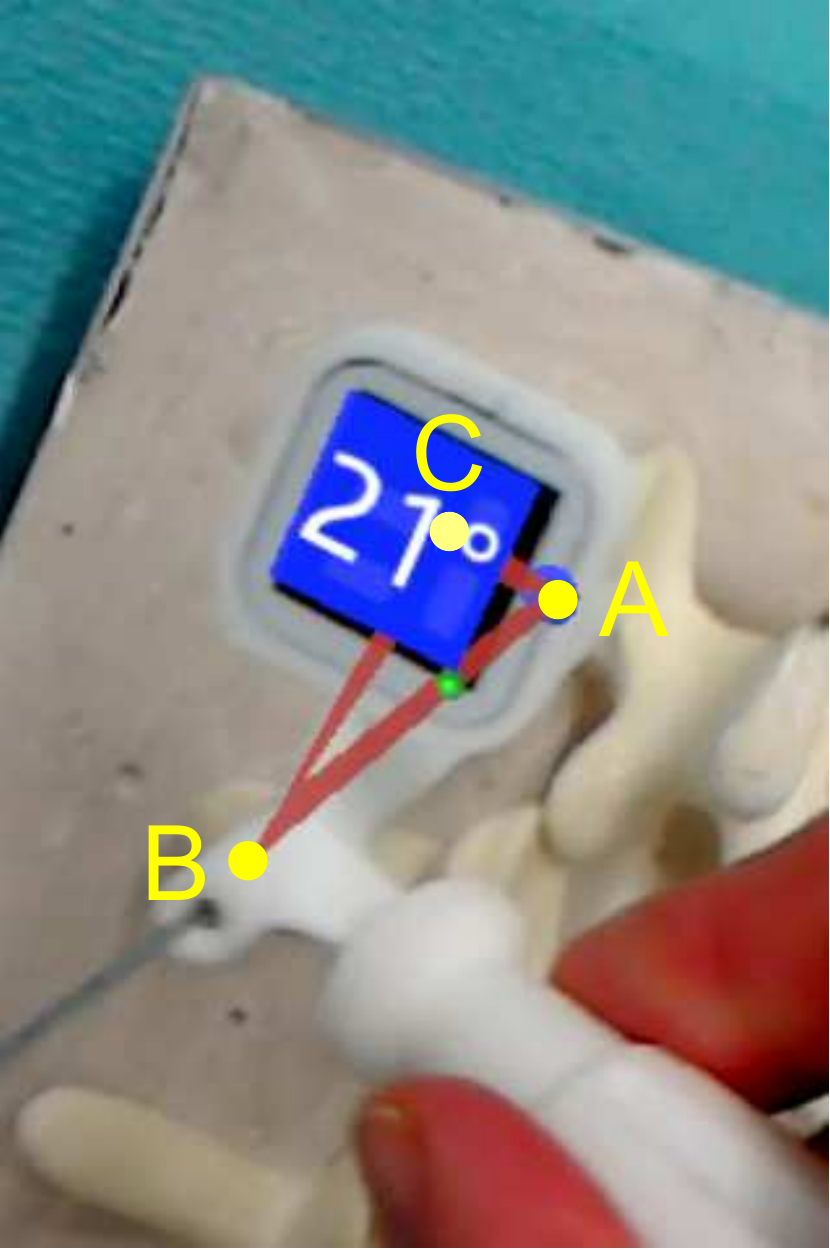}
			\caption{}
			\label{fig5d}
		\end{subfigure}
	\end{center}
	\caption{a) A surgeon uses the custom-made surgical navigation tool. b) Entry point overlay (blue) in the beginning of the navigation. c) Augmented view of the surgeon during navigation (except yellow points).
		\vspace{2mm}\newline\noindent
		\tiny
		\textit{Reprinted by permission from RightsLink: Springer International Journal of Computer Assisted Radiology and Surgery \cite{liebmann2019pedicle}, \textcopyright\space CARS (2019), advance online publication, 15 April 2019 (doi: \href{https://doi.org/10.1007/s11548-019-01973-7}{10.1007/s11548-019-01973-7}.IJCARS.)}
	}
	\label{fig5}
\end{figure}

Screw navigation relies on a trackable (Section \ref{subsec:marker_tracking}) custom-made surgical navigation tool (NT, Figures \ref{fig5b} and \ref{fig5d}). Handle and sleeve are designed such that they allow holding the ND with one hand, while inserting a K-wire with the other (Figure \ref{fig5b}).
Navigation can be started after successful registration (Section \ref{subsec:registration}). First, the screw entry points are shown (Figure \ref{fig5c}). The pointed K-wire tip is pierced into the bone at the targeted position, purely relying on holographic visualization. The K-wire tip does not slide away and the surgeon can start to navigate towards the desired trajectory. Thereby, holographic feedback comprises two parts of information (Figure \ref{fig5d}). First, the 3D angle between current and targeted trajectory is displayed. Second, a triangle is rendered between three virtual points: the screw entry point (A), a point lying on the current trajectory of the NT (B) and one lying on the targeted screw trajectory (C). This way, the surgeon is given an intuitive feedback about trajectory deviation in 3D space.

\begin{figure}[htb]
	\begin{center}
		\begin{subfigure}{0.49\linewidth}
			\includegraphics[width=\linewidth]{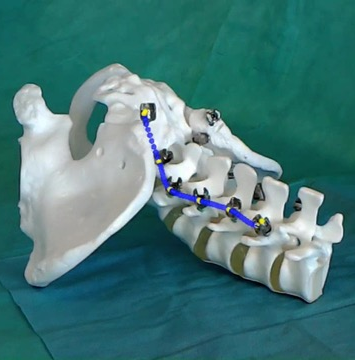}
			\caption{}
			\label{fig8a}
		\end{subfigure}
		\begin{subfigure}{0.49\linewidth}
			\includegraphics[width=\linewidth]{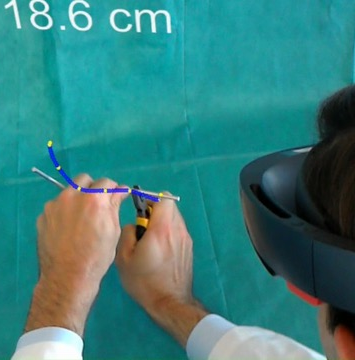}
			\caption{}
			\label{fig8b}
		\end{subfigure}
	\end{center}
	\caption{a) The rod template. b) Shared experience capture of a surgeon bending a rod according to the template. Please note that both images originate from \cite{wanivenhaus2019augmented}.}
	\label{fig8}
\end{figure}

Rod bending navigation is based on calculating and displaying a holographic template of the optimal rod shape. It does not require registration with the anatomy, but the same tracked PD (Figure \ref{fig2a}) is employed to capture the head position for each screw. From these 3D points, a centripetal Catmull-Rom spline is calculated and the rod length is estimated (Figure \ref{fig8a}). A Catmull-Rom spline is a piecewise function that passes through its defining points \cite{catmull1974class}, i.e. the stored screw head positions. We use the centripetal parametrization as iof self-intersections and cusps within curve segments \cite{yuksel2009parameterization}. Finally, the newly created spline is moved away from the patient and the rod is bent according to the holographic template (Figure \ref{fig8b}).

\subsection{Pre-clinical evaluation}
\label{subsec:evaluation}
\begin{figure}[htb]
	\begin{center}
		\begin{subfigure}{0.32\linewidth}
			\includegraphics[width=\linewidth]{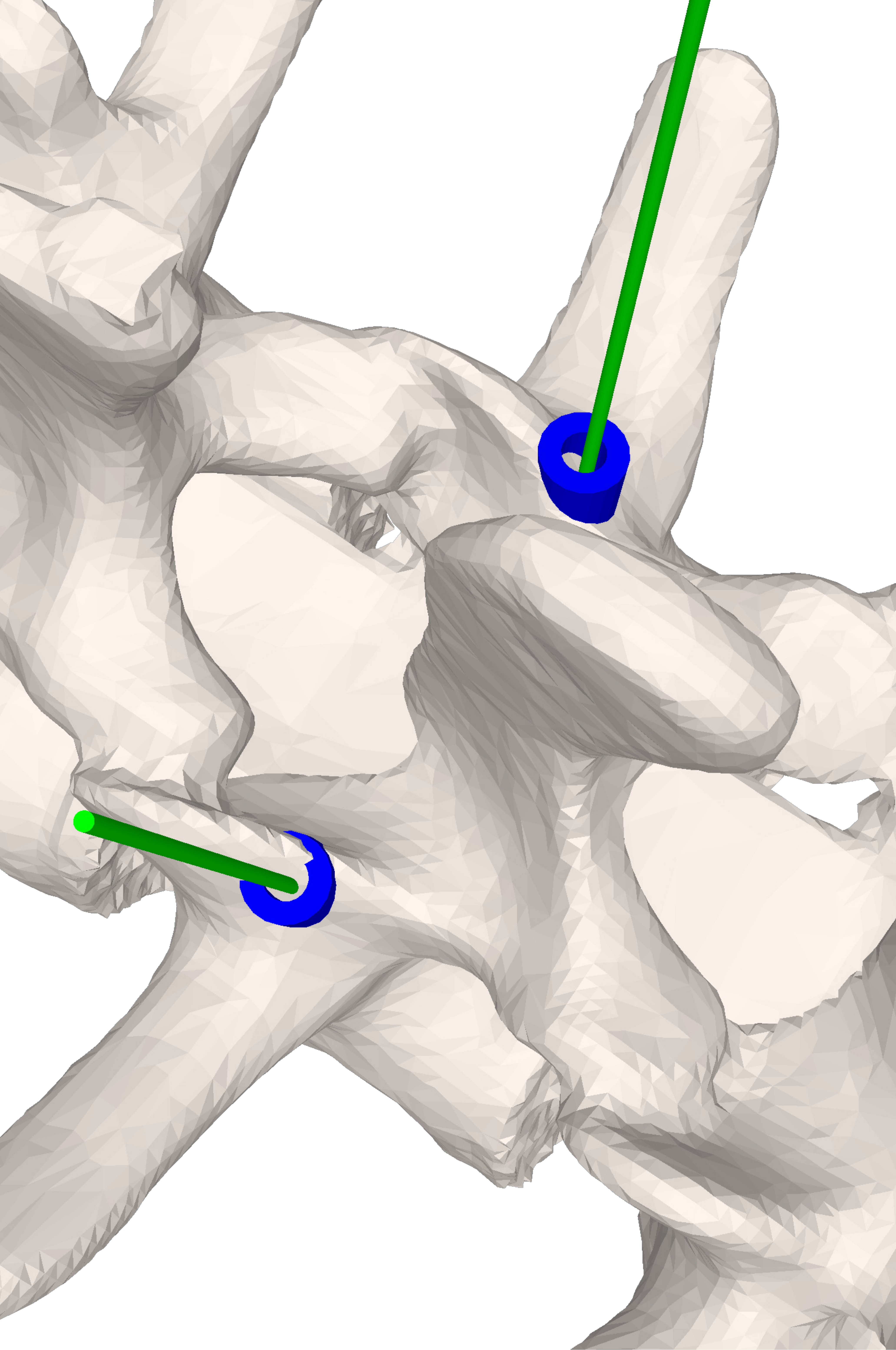}
			\caption{}
			\label{fig6a}
		\end{subfigure}
		\begin{subfigure}{0.32\linewidth}
			\includegraphics[width=\linewidth]{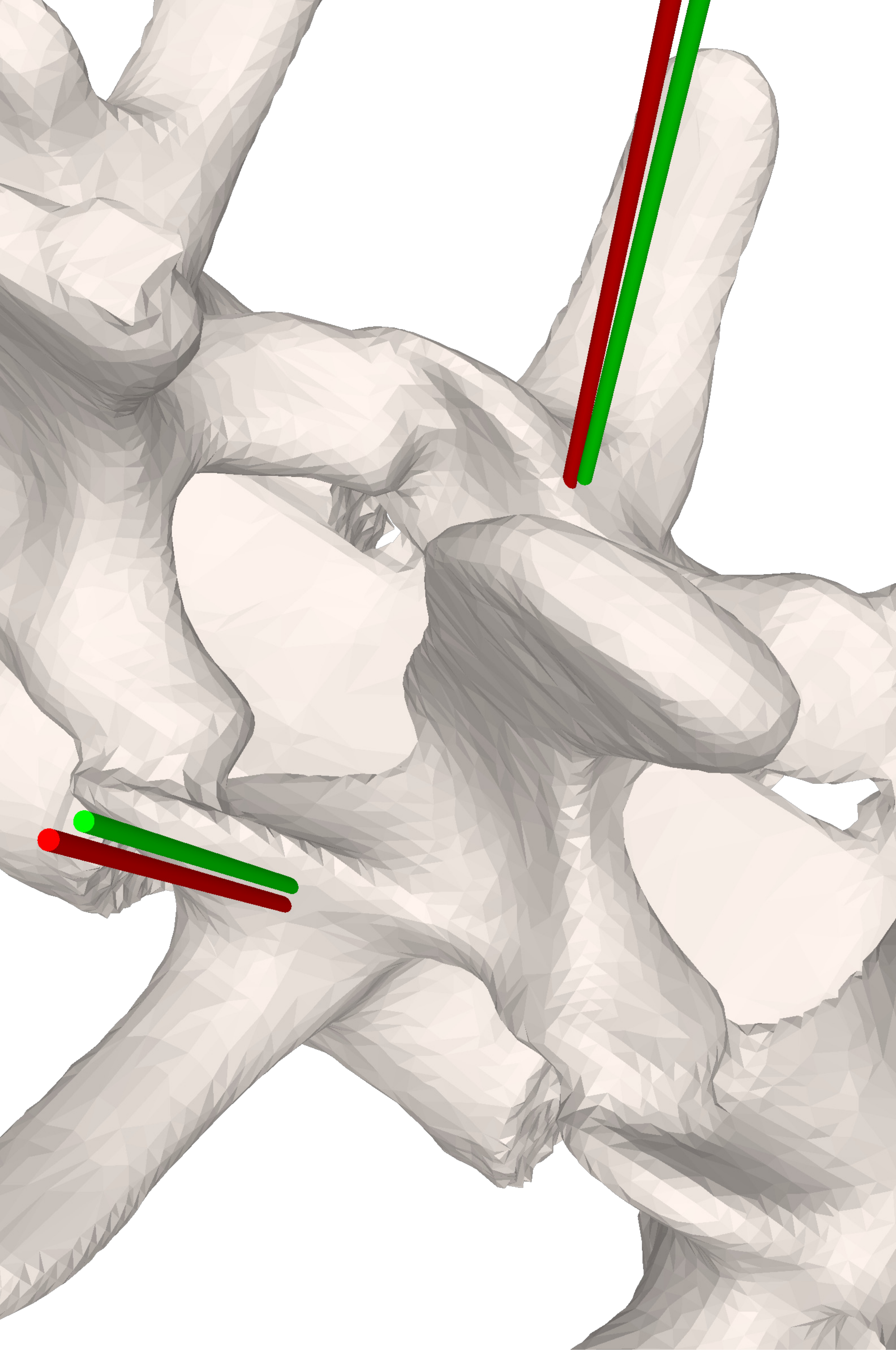}
			\caption{}
			\label{fig6c}
		\end{subfigure}
		\begin{subfigure}{0.32\linewidth}
			\includegraphics[width=\linewidth]{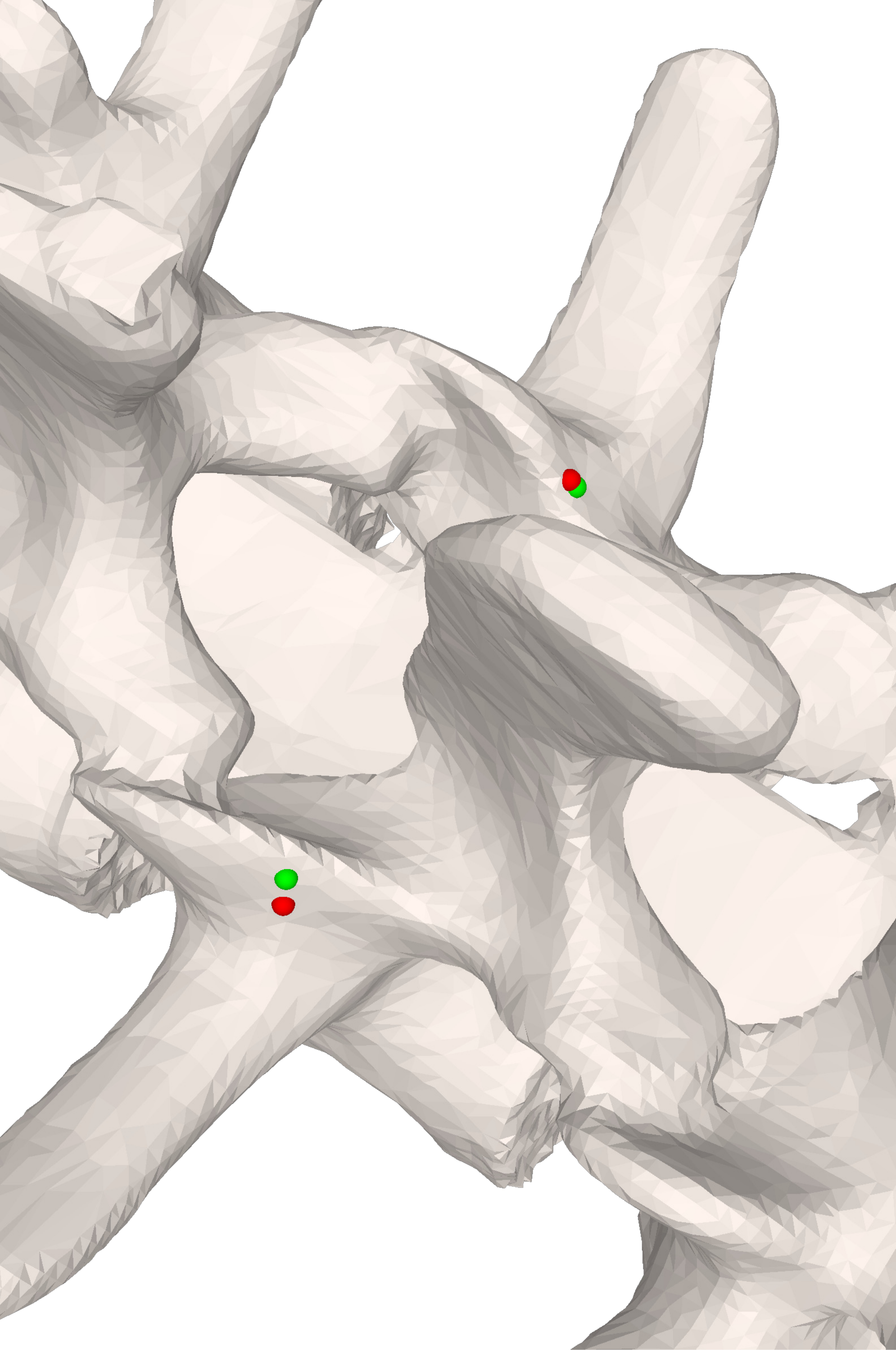}
			\caption{}
			\label{fig6d}
		\end{subfigure}
	\end{center}
	\caption{a) Planned trajectories (green) and entry points (blue). b) Planned trajectories (green) and red cylinder representing postoperatively segmented K-wires. c) Planned (green) and postoperatively quantified entry points (red).
		\vspace{2mm}\newline\noindent
		\tiny
		\textit{Reprinted by permission from RightsLink: Springer International Journal of Computer Assisted Radiology and Surgery \cite{liebmann2019pedicle}, \textcopyright\space CARS (2019), advance online publication, 15 April 2019 (doi: \href{https://doi.org/10.1007/s11548-019-01973-7}{10.1007/s11548-019-01973-7}.IJCARS.)}
	}
	\label{fig6}
\end{figure}

Our method for screw navigation was evaluated on two phantoms of the lower lumbar spine that were molded into a plastic tub to mimic clinical accessibility of vertebra surfaces (e.g. Figure \ref{fig5c}). Screw trajectories were planned preoperatively (Figure \ref{fig6a}). Each vertebra was registered individually and two K-wires (left and right) representing the pedicle screws were inserted, solely relying on AR navigation. After execution, postoperative 3D models of vertebrae and inserted K-Wires were generated from CT scans and aligned to the preoperative plan. For each K-wire, trajectory and entry point were quantified. The trajectory was defined by aligning a generic cylindrical object to the segmented K-wire (Figure \ref{fig6c}). The entry point was defined as the first point along the trajectory intersecting with the preoperative 3D model (Figure \ref{fig6d}). Primary results were the 3D angle between planned and executed trajectories as well as the 3D distance between planned and executed entry points. Secondary results for each vertebra included registration error, surface digitization time and the number of sampled points.

The rod bending navigation was evaluated on an instrumented phantom of the lumbar spine (Figure \ref{fig8a}) by comparison with the standard trial-and-error technique. For each group, a total of six rods were bent by three surgeons. The time for bending and insertion, the number of rod-rebending maneuvers and the rod length accuracy were recorded.

\section{Results and discussion}
Results of our screw navigation experiments are listed in Table \ref{table1}. The mean registration RMSE (1.62 mm) and time (125 s) are comparable to state-of-the-art systems (e.g. 0.9 mm and 117 s in Nottmeier and Crosby \cite{nottmeier2007timing}). The accuracy is promising and suggests that the method has the potential for clinical use.
\begin{table}[htb]
	\begin{center}
		\begin{tabular}{@{}lcccc@{}}
			\toprule
			Result	& Mean & SD & min. & max. \\
			\midrule
		 	Trajectory err. (\degree)	 & 3.38 & 1.73 & 1.16&6.33 \\
			Entry point err. (mm)	& 2.77  & 1.46 & 0.71 & 7.20\\
			\midrule
			Reg. RMSE (mm) & 1.62& 0.26& 1.14& 2.02\\
			Digitization time (s) & 125& 27& 91& 185\\
			\# points collected & 1’983 & 404 & 1’268 & 2’744\\
			\bottomrule
		\end{tabular}
	\end{center}
	\caption{Results of screw navigation experiments.}
	\label{table1}
\end{table}

All of the rod bending navigation results were in favor of the AR group, with time exposure (374 vs. 465 s) and rod length (15/18 vs 4/18 correct) showing significant differences. Clinical trials are imminent.

The screw navigation method has limitations. Our experimental setup was simplified by rigidly attaching the phantoms to a table. This obviated motion compensation, but holograms are prone to drift after placement \cite{vassallo2017hologram}. This may have negatively influenced our results, although the surgeon tried to minimize head movement during the experiments. On another note, the use of Research Mode sensors for stereo vision negatively affects application stability as they are an essential part of the HoloLens' built-in SLAM. It is assumed that all such capabilities will perform even better in upcoming releases.

Our results encourage 3D/3D image-to-patient registration. For future work, we plan to employ RGB-D sources, such as Azure Kinect DK, promoting an automated process.

{\small
\bibliographystyle{ieee_fullname}
\bibliography{registration}
}

\end{document}